\documentclass[pdflatex,sn-nature]{sn-jnl}% Style for submissions to Nature Portfolio journals

\usepackage{graphicx}%
\usepackage{float}%
\usepackage{multirow}%
\usepackage{amsmath,amssymb,amsfonts}%
\usepackage{amsthm}%
\usepackage{mathrsfs}%
\usepackage[title]{appendix}%
\usepackage{xcolor}%
\usepackage{textcomp}%
\usepackage{manyfoot}%
\usepackage{booktabs}%
\usepackage{algorithm}%
\usepackage{algorithmicx}%
\usepackage{algpseudocode}%
\usepackage{listings}%
\usepackage{soul}
\usepackage{verbatim}
\usepackage{lipsum} 
\usepackage[numbers,sort&compress]{natbib}

\usepackage{blindtext}
\usepackage{geometry}
\begin{document}

\title[Article Title]{
GLU: Global-Local-Uncertainty Fusion for Scalable Spatiotemporal Reconstruction and Forecasting
}

\author*[1]{\fnm{Linzheng} \sur{Wang}}\email{wanglz@mit.edu}

\author[1]{\fnm{Jason} \sur{Chen}}\email{jachen25@mit.edu}

\author[1]{\fnm{Nicolas} \sur{Tricard}}\email{ntricard@mit.edu}

\author[1]{\fnm{Zituo} \sur{Chen}}\email{zituo@mit.edu}

\author*[1]{\fnm{Sili} \sur{Deng}}\email{silideng@mit.edu}

\affil[1]{\orgdiv{\textit{Department of Mechanical Engineering}}, \orgname{\textit{Massachusetts Institute of Technology}}, \orgaddress{\city{\textit{Cambridge}}, \postcode{\textit{02139}}, \state{\textit{MA}}, \country{\textit{United States of America}}}}

\abstract{
Digital twins of complex physical systems are expected to infer unobserved states from sparse measurements and predict their evolution in time, yet these two functions are typically treated as separate tasks. 
Here we present GLU, a Global–Local–Uncertainty framework that formulates sparse reconstruction and dynamic forecasting as a unified state-representation problem and introduces a structured latent assembly to both tasks.
The central idea is to build a structured latent state that combines a global summary of system-level organization, local tokens anchored to available measurements, and an uncertainty-driven importance field that weights observations according to the physical informativeness. 
For reconstruction, GLU uses importance-aware adaptive neighborhood selection to retrieve locally relevant information while preserving global consistency and allowing flexible query resolution on arbitrary geometries. 
Across a suite of challenging benchmarks,
GLU consistently improves reconstruction fidelity over reduced-order, convolutional, neural-operator, and attention-based baselines, better preserving multi-scale structures. 
For forecasting, a hierarchical Leader–Follower Dynamics module evolves the latent state with substantially reduced memory growth, maintains stable rollout behavior and delays error accumulation in nonlinear dynamics. 
On a realistic turbulent combustion dataset, it further preserves not only sharp fronts and broadband structures in multiple physical fields, but also their cross-channel thermo-chemical couplings. 
Scalability tests show that these gains are achieved with substantially lower memory growth than comparable attention-based baselines. 
Together, these results establish GLU as a flexible and computationally practical paradigm for sparse digital twins.
} 


\maketitle
\newpage

%\section{Introduction}\label{sec1}

\noindent Digital twins have the potential to transform the design, monitoring, and proactive control of complex scientific and engineering systems by continuously updating virtual replicas of physical assets~\cite{tao2024advancements}.
While physical systems evolve continuously in time and space, available measurements are often sparse, intermittent, and noisy~\cite{singh2021digital}.
From turbulent flows~\cite{vinuesa2024perspectives} to reactive combustion~\cite{wang20243} and geophysical climate systems~\cite{wang2025vast}, the ability to both infer unobserved system states from limited measurements and predict their evolution is a prerequisite for intelligent autonomy~\cite{piccialli2025digital, DENG2026140506, deng2025scientific,san2026evolution}.

Current approaches predominantly treat sparse reconstruction and spatiotemporal forecasting as distinct tasks, and therefore rarely pursue a state representation explicitly designed to support both accurate sparse inference and stable dynamics learning.
Reconstruction models typically optimize snapshot fidelity without explicitly designing representations for downstream temporal evolution~\cite{fukami2021global, santos2023development, liu2025enhancing, WANG2025}; while forecasting models generally assume the availability of dense, high-quality initial states~\cite{SoleraRico2024, kontolati2024learning, tomasetto2025reduced}, which are rarely afforded in practice.
This decoupling creates a critical vulnerability. 
Errors in the initial state estimation, arising from the ill-posed nature of sparse reconstruction, typically propagate through the forecasting model, leading to unphysical trajectories and rapidly degrading long-horizon accuracy~\cite{brajard2020combining, chattopadhyay2020data, cheng2024efficient}.
Robust digital twins therefore require reconstruction and forecasting to be built around a common state representation~\cite{Huang2024representation}, rather than as loosely coupled sequential modules.
% This problem is also closely related to data assimilation, which combines observations with predictive models to estimate system state and is increasingly viewed as a core ingredient of digital twins~\cite{Li2023Big}. 
% However, most assimilation frameworks assume either an explicit physical forecast model or a dedicated update procedure, whereas our goal here is different: to design a learned sparse-observation state representation that supports both full-state inference and forecasting within a single computational backbone.
Recent work has highlighted the importance of representation quality in spatiotemporal physical systems, showing that latent predictive objectives can better support downstream tasks than direct pixel-level prediction~\cite{qu2026representation}.

The key difficulty is that this state must be simultaneously inferable from sparse observations, expressive enough to preserve local physical detail, and structured enough to support stable temporal evolution.
% The central challenge in this unification is to overcome the information bottleneck inherent in representing physical states from limited observation data.
Existing methods face a dichotomy between local and global paradigms.
In effect, sparse digital twins require a state assembly that preserves local measurement fidelity without collapsing long-range physical organization into an overly compressed bottleneck.
Local information-preserving models, such as graph networks~\cite{ogoke2021graph} or convolutional inpainting~\cite{zhang2023improved}, excel at preserving fine-scale details near sensors but struggle to capture long-range correlations or global modes essential for dynamics~\cite{10.5555/3618408.3618720}.
Conversely, global representation models, such as reduced-order models (ROM) or Neural Operators, encode observations into compact latent states to capture system-level physics~\cite{callaham2019robust, bahmani2025resolution, Xia_2025_CVPR}.
However, such compression often discards sensor-specific fine-scale information, resulting in a fidelity gap where local anomalies are smoothed out~\cite{KHODAKARAMI2026108027}.

\begin{figure}[htbp]
    \centering
    \includegraphics[width=0.95\linewidth]{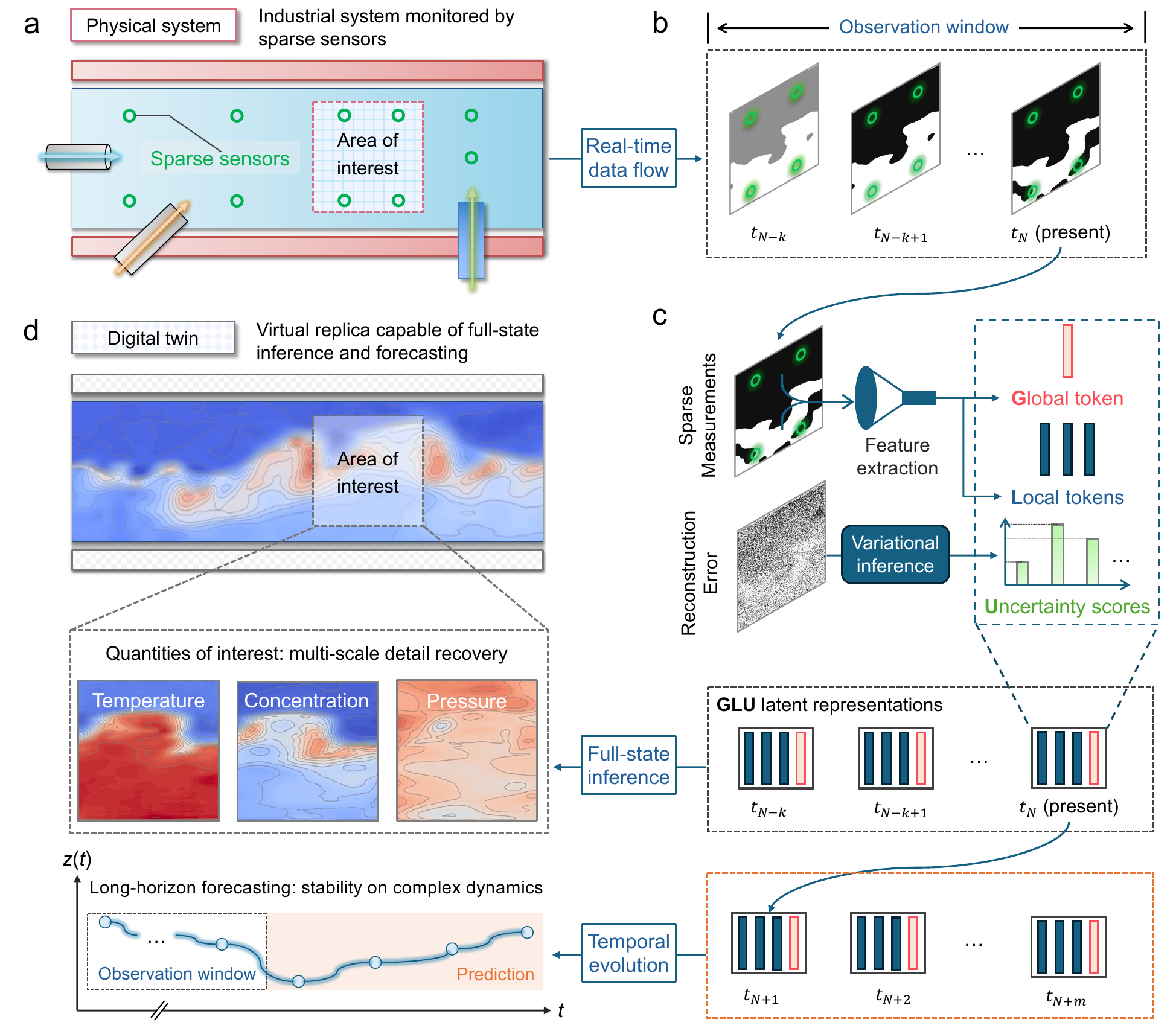}
    \caption{ \textbf{Bridging the reality gap between sparse sensing and dynamic digital twins.}
    a) Complex physical systems monitored by sparse, irregular sensors. 
    b) A continuous stream of sparse, partial observations serves as the input. 
    c) The GLU framework decomposes measurements into three complementary streams. 
    Global tokens extract system-level patterns and long-range correlations, while local tokens preserve fine-grained sensor fidelity. 
    A learned uncertainty signal guides the assimilation of local data. 
    d) The GLU-assembled representations drive the digital twin, enabling both full-state inference of unobserved quantities and stable forecasting of future trajectories.}
    \label{fig:idea_overview}
\end{figure}

In a unified reconstruction and forecasting pipeline, the goal is to learn a physically meaningful state representation that can be robustly inferred from arbitrary sparse measurements and can be faithfully evolved forward in time~\cite{shen2022nerp, NEURIPS2022_cf70320e, cai2025causal}. 
Most existing approaches offer only partial solutions to this joint problem.
In sparse regimes, where reconstruction uncertainties arise from both sensor noise and underdetermined measurements~\cite{der2009aleatory}, designing effective state representations must balance scalability with retrieval accuracy.
Addressing this challenge therefore requires adherence to two principles: \textit{information maximization}, ensuring that every available measurement can be effectively retained and utilized; and \textit{information weighting}, recognizing that information density is non-uniform in space and time~\cite{zhong2024novel}, thereby prioritizing data from dynamically informative regions while explicitly managing uncertainty in under-sampled areas.

In this work, we introduce the Global–Local–Uncertainty (GLU) framework, a unified architecture to integrate sparse information streams for high-fidelity state retrieval and dynamic forecasting (\textbf{Figure~\ref{fig:idea_overview}}), bridging the gap between sparse sensing and dynamic digital twins.
GLU explicitly fuses three complementary streams of information. 
A global stream encodes a compact, system-level latent representation that captures long-range correlations and larger pattern structure. 
A local stream preserves tokenized features anchored to specific sensor locations, mitigating information loss. 
An uncertainty stream provides a learned, spatially resolved importance score that gates information flow, allowing the model to prioritize reliable data.

We evaluate GLU across a suite of challenging benchmarks, including periodic and chaotic flows~\cite{fukami2021global, SoleraRico2024}, reaction–diffusion systems~\cite{Rao2023Encoding}, and a realistic multi-physics turbulent combustion dataset. 
GLU consistently outperforms specialized reconstruction and forecasting baselines, achieving state-of-the-art accuracy and efficiency in both full-field metrics and physically meaningful quantities, while preserving multi-scale structure and cross-channel physical couplings. 
Our method provides a powerful foundation for sparse adaptation in next-generation digital twins, bringing data-driven state estimation and forecasting a step closer to practical deployment.

\section*{Results}
\label{sec: Results}

\subsection*{Global-local-uncertainty information fusion}
\label{Subsec: ModelOutline}

\begin{figure}[htbp]
    \centering
    \includegraphics[width=\linewidth]{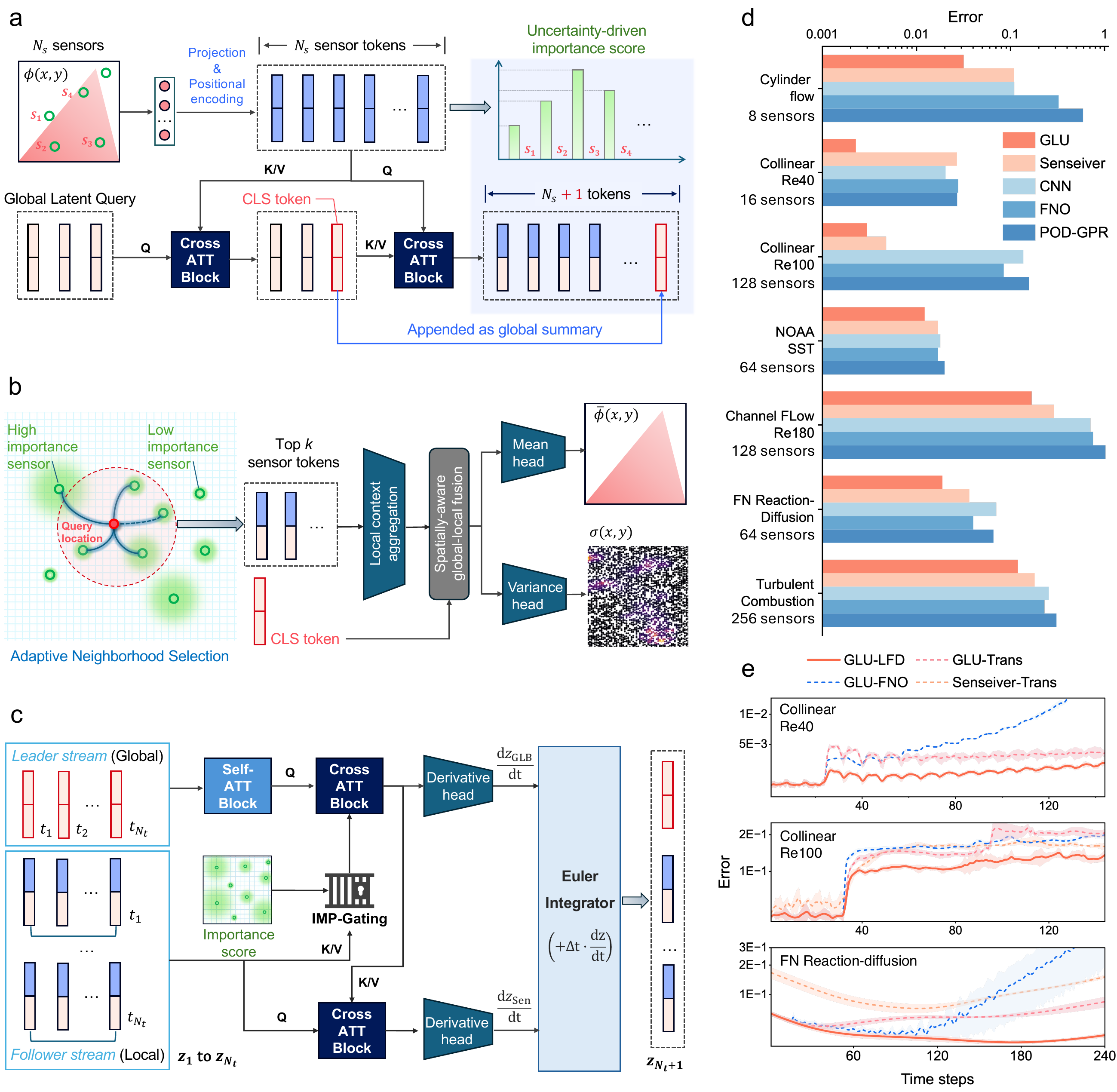}    \caption{\textbf{Architecture and performance of the GLU framework.} 
    \textbf{a,} Sparse encoding via global-local-aware encoder with bidirectional attention to transform arbitrary sparse inputs into a latent state.
    % , consisting of $N_s$ locally-anchored sensor tokens (blue) and a single global leader token (red). 
    \textbf{b,} Soft adaptive reconstruction. For any query location, the model dynamically aggregates top-$k$ most relevant sensors using an importance-warped distance metric. 
    This process decodes both the mean field value and a spatial uncertainty estimate. 
    \textbf{c,} Leader-Follower-Dynamic structure to model latent temporal evolution.
    % Temporal evolution is modeled as a Neural ODE with a Leader-Follower-Dynamic structure. 
    The global leader drives the dynamics by attending to important regions via IMP-Gating, while sensor followers update their local states by attending to the leader. 
    \textbf{d,} Normalized reconstruction error across seven datasets ranging from canonical flows to chaotic reaction-diffusion and multi-physics turbulent combustion. 
    % GLU consistently outperforms leading baseline methods. 
    \textbf{e,} Forecasting error accumulation for periodic (Collinear Re40) and chaotic (Collinear Re100, Reaction-Diffusion) dynamics. 
    % GLU significantly suppresses error propagation compared to standard transformer or neural operator baselines.
    }
    \label{fig:model_overview}
\end{figure}

As illustrated in Figure~\ref{fig:model_overview}, the GLU framework represents sparse physical states through coupled global, local, and importance-weighted components.
Consider a physical system observed via a sparse set of $N$ sensors at locations $\mathbf{x} = \{x_i\}_{i=1}^N$ with values $\mathbf{u} = \{u_i\}_{i=1}^N$. 
We represent the latent state of the system at time $t$ as:
\begin{equation}
    \mathcal{S}_t = \{ \mathbf{z}_{\text{global}}^{(t)}, \ \mathbf{Z}_{\text{local}}^{(t)} \},
\end{equation}
where $\mathbf{z}_{\text{global}} \in \mathbb{R}^{D}$ is a single \textit{Leader} (CLS) token capturing the holistic system state, $\mathbf{Z}_{\text{local}} = \{z_i\}_{i=1}^N \in \mathbb{R}^{N \times D}$ corresponds to the set of $N$ \textit{Follower} sensor tokens preserving local details.
In parallel, each sensor token is associated with an importance score $\boldsymbol{\phi} \in [0,1]^N$, which quantifies the inferred information value of the available measurements under sparse observability.

Standard tokenization (e.g., in Vision Transformers~\cite{dosovitskiy2021an} or Perceiver~\cite{pmlr-v139-jaegle21a}) often discards local details or imposes a bottleneck.
This is acceptable under information redundant regime (with access to full field), but can impose strong limitation in sparse regime.
To achieve information-preserving encoding on sparse data, we introduce a Global-local-aware encoder with bidirectional flow, as shown in \textbf{Figure~\ref{fig:model_overview}a}.
First, a small set of learnable latent queries cross-attends to the raw sensor inputs to extract a global summary. 
Crucially, instead of discarding the inputs, we inject this global context back into the $N$ sensor tokens via a second cross-attention layer. 
This enables every sensor token $z_i$ to encode not only its local measurement $u_i$ but also its relationship to the global field, with linear complexity in the number of sensors. 
The final encoded state includes the enriched sensor tokens and the appended global leader token.

To infer the field value at an arbitrary query location $y$, we propose a soft, importance-aware, domain decomposition strategy (\textbf{Figure~\ref{fig:model_overview}b}). 
Instead of fixed geometric partitions like Voronoi, we define an adaptive neighborhood by an information--aware metric. 
The relevance of a sensor $i$ to query $y$ is determined by a weighted distance $d_{\phi}(y, x_i)$, which fuses Euclidean proximity with the learned importance score $\phi_i$:
\begin{equation}
    \hat{u}(y) = \mathcal{D}_{\theta}\left( \text{kNN}_{\phi}(y, \mathbf{Z}_{\text{local}}), \mathbf{z}_{\text{global}} \right)
\end{equation}
Here, $\text{kNN}_{\phi}$ selects the top-$K$ sensors that maximize information utility, allowing highly important sensors to influence larger regions. 
The decoder $\mathcal{D}_{\theta}$ fuses these selected local tokens with the global token to predict the value, ensuring predictions are locally grounded yet globally consistent.

The spatially-aware fusion block decodes not only the mean field value but also a pixel-wise variance estimate using a heteroscedastic neural network~\cite{Xia_2025_CVPR}.
% The variance estimates uncertainty in the decoded field, and then serves as the supervision signal for learning the importance scores $\boldsymbol{\phi}$.
This predicted uncertainty is then used to supervise the importance scores $\boldsymbol{\phi} $ through a variational auxiliary objective, encouraging the model to upweight sensors that are most informative for resolving uncertain regions.
Technical details of this variational formulation are demonstrated in \textbf{Methods}.

Evolving the full set of local tokens directly is both computationally expensive and statistically inefficient because most pairwise interactions are not equally informative.
We resolve this via a hierarchical Leader-Follower-Dynamics (LFD) architecture (\textbf{Figure~\ref{fig:model_overview}c}).
The global leader token $\mathbf{z}_{\text{global}}$ acts as the driver of the dynamics. 
It evolves by integrating its recent history while aggregating information from the local tokens.
This aggregation is weighted by the importance scores $\boldsymbol{\phi}$, focusing the model's attention on dynamically significant regions like wake vortices or reaction fronts:
\begin{equation}
    \mathbf{z}_{\text{global}}^{(t+1)} \leftarrow \text{Attention}(\mathbf{z}_{\text{global}}^{(1:t)}, \mathbf{Z}_{\text{local}}^{(t)}; \boldsymbol{\phi}).
\end{equation}
In practice, this weighting can be implemented as an additive bias on the attention logits, so that sensors with $\phi \rightarrow 0$ contribute negligibly to the leader update, whereas high-importance sensors are preferentially attended.
Conversely, the local sensor tokens $\mathbf{Z}_{\text{local}}$ act as followers. 
They do not model dense pairwise interactions, but instead evolve by attending to the leader's trajectory, effectively receiving updated global context from the leader.
This hierarchical design largely reduces forecasting complexity from dense pairwise token interactions to approximately linear scaling in the number of sensors.
% Additional architectural details, hyperparameters, and training protocols for GLU and all baselines are provided in the Supplementary Information (SI).

To demonstrate the broad applicability of GLU, we assess its performance across a hierarchy of physical complexity, ranging from canonical fluid dynamics (cylinder flow, collinear plate flow), NOAA sea surface temperature (SST), to chaotic turbulent channel flow and FitzHugh-Nagumo reaction–diffusion systems, and finally to a realistic multi-physics turbulent combustion dataset. 
Throughout the paper, model performance is evaluated using the relative $L_2$ error.
As summarized in \textbf{Figure~\ref{fig:model_overview}d}, GLU consistently achieves superior reconstruction accuracy compared to state-of-the-art baselines (e.g., Senseiver, CNN, POD-GPR, FNO).
Furthermore, in long-term forecasting tasks using sparse observation as input(\textbf{Figure~\ref{fig:model_overview}e}), the hierarchical LFD structure significantly suppresses error accumulation, maintaining stability over long horizons even for chaotic systems where traditional baselines diverge.
GLU's advantage is consistent across both smooth large-scale fields and strongly multi-scale systems, indicating that the fused representation generalizes well across very different physical systems.

\subsection*{High-fidelity adaptive spatial reconstruction}
\label{Subsec: ReconPerformance}

\begin{figure}[!htbp]
    \centering    \includegraphics[width=\linewidth]{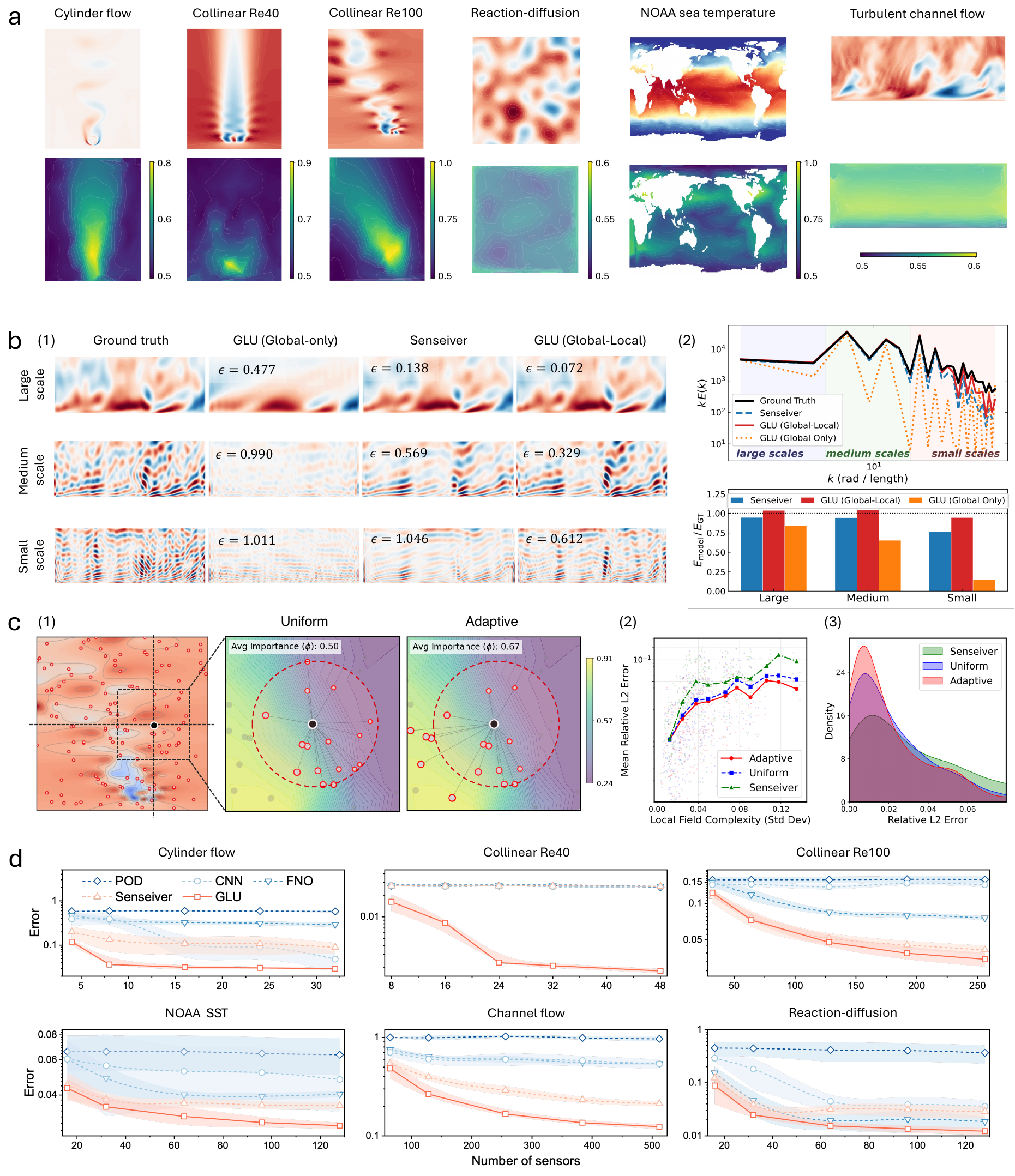}
    \caption{\textbf{Adaptive spatial reconstruction mechanisms and information scaling performances.} 
    \textbf{a,} Learned importance score distributions ($\phi$) for representative datasets. 
    % The model automatically highlights physically complex regions, such as wakes, shear layers, and reaction fronts, without explicit supervision.
    \textbf{b,} Multi-scale detail recovery:
    (1) Ground truth and scale-separated reconstructions (large/medium/small) from global-only ablation, Senseiver, and GLU; 
    % $\varepsilon$ denotes scale-wise relative $L_2$ error.
    % CLS-only captures the dominant large-scale structure but suppresses fine-scale fluctuations; Senseiver retains localized variability with reduced coherence at higher $k$.
    % GLU is most accurate across scales, with the largest gains at small scales.
    (2) Spectral and scale-wise summaries. 
    GLU best matches the ground-truth energy spectrum and retains the most energy across scales (bar chart), yielding the lowest error at each scale.
    % Comparison of reconstruction fidelity on Turbulent Channel Flow. The Global-only ablation (left) captures mean structure but over-smooths turbulence; 
    % Senseiver-style latent array captures some details but lacks coherence; 
    % GLU fuses both to recover fine-scale turbulent fluctuations. 
    \textbf{c,} Mechanism of adaptive selection:
    (1) Visualization of the adaptive neighborhood on Collinear Flow; 
    % Unlike uniform selection (black circle), importance-weighting (warped lines) allows the query point to attend to distant but informative sensors (red circles). 
    (2) Reconstruction error vs. local field complexity; 
    % GLU maintains robustness in complex regions where uniform selection degrades. 
    (3) Probability density function of reconstruction errors;
    % , showing GLU shifts the distribution toward zero error.
    \textbf{d,} Reconstruction error versus number of sensors across six datasets. 
    % showing improved information efficiency as more observations become available. 
    } \label{fig:uncertainty_spatial}
\end{figure}

GLU improves spatial reconstruction by learning where observations are most informative and combining that information with a representation that remains both globally coherent and locally faithful.
As visualized in \textbf{Figure~\ref{fig:uncertainty_spatial}a}, without explicit spatial supervision, the model learns to assign higher importance to dynamically active regions.

In separated flows, like cylinder flow and collinear flow, the importance scores concentrate around wakes and shear layers.
For the NOAA SST dataset, the importance distribution is concentrated along coastlines and major boundary currents, highlighting regions dominated by strong seasonal variability and advective transport while assigning lower importance to the comparatively stable open ocean. 
In contrast, for systems characterized by homogeneous isotropic dynamics, including the reaction-diffusion system and turbulent channel flow, the learned importance scores become nearly uniform across the domain. 
In these cases, no localized region consistently dominates the dynamics, and the model correctly infers that the information is distributed broadly across space. 
The emergence of physically interpretable importance patterns without explicit supervision indicates that the proposed uncertainty-driven weighting mechanism learns to identify where observations are most informative for state estimation.

To further understand the structural advantages of GLU, we isolate the contributions of its representational components in recovering multi-scale turbulence, with particular emphasis on small-scale structure in \textbf{Figure~\ref{fig:uncertainty_spatial}b}. 
We compare GLU against two contrasting representational paradigms on the turbulent channel flow dataset: a global-only model, analogous to reduced-order modeling, and Senseiver, which is based on a fixed set of tokens. 
The global-only model captures the dominant large-scale energy distribution but behaves as a strong low-pass filter, leading to a pronounced loss of high-wavenumber energy and the suppression of fine-scale turbulent fluctuations. 
Senseiver partially alleviates this limitation by retaining localized variability, but still exhibits a systematic deficit in high-wavenumber energy and reduced fidelity across intermediate scales. 
Spectral analysis in n \textbf{Figure~\ref{fig:uncertainty_spatial}b}-(2) confirms that GLU achieves the closest agreement with the ground-truth energy spectrum across the full wavenumber range, with the most pronounced improvements occurring in the small-scale regime, where competing methods exhibit substantial energy loss. 
Consistent with this observation, scale-resolved error analysis shows that GLU provides the most accurate reconstruction across large, medium, and small spatial scales.

The importance-guided adaptive neighborhood selection is also the key mechanism driving reconstruction fidelity.
As shown in \textbf{Figure~\ref{fig:uncertainty_spatial}c}-(1), in standard uniform selection (e.g., $k$-NN with Euclidean distance), the reconstruction is limited by a fixed geometric radius, forcing the model to rely on nearby sensors even if they are uninformative. 
GLU utilizes the learned importance scores to warp the search metric, effectively expanding its receptive field to recruit high--importance sensors from further away when the local neighborhood is sparse or noisy. 
This is clearly visible in the selection visualization, where the query point bypasses closer, low-importance sensors to attend to distant, high-information anchors. 
This adaptivity confers robustness. 
In \textbf{Figure~\ref{fig:uncertainty_spatial}c}-(2), we show that as local field complexity increases, the error for uniform selection rises sharply, whereas the adaptive selection maintains a flatter error profile. 
Statistically, this results in a probability density distribution of reconstruction errors that is significantly more peaked around zero in \textbf{Figure~\ref{fig:uncertainty_spatial}c}-(3), confirming that adaptive selection effectively eliminates the heavy tail of large errors typically found in complex flow regions.
Together, these results show that GLU has superior performance because it preserves the multi-scale structure for faithful reconstruction and allocates representational capacity where the field is hardest to infer.

Furthermore, the multi-scale information preservation and importance-aware processing designs of GLU are directly linked to the model's superior information scaling capability, as quantified in \textbf{Figure~\ref{fig:uncertainty_spatial}d}. 
We evaluate reconstruction error as a function of sensor density across six diverse physical systems without retraining the model. 
In the ultra-sparse regime (left side), all methods are fundamentally limited by a lack of information, and GLU's performance is comparable to leading baselines like Senseiver. 
However, as the sensor density increases, GLU demonstrates a significantly faster rate of error reduction compared to all other methods. While baselines like CNN and FNO quickly hit a performance plateau, GLU continuously and efficiently integrates the additional sensor data to refine its reconstruction. 
This ability to scale up performance dramatically with available information, without structural changes, highlights the effectiveness of its adaptive fusion mechanism.

\subsection*{Long-term forecasting stability from sparse observations}
\label{Subsec: ForecastingPerformance}

Robust digital twins require forecasting models that remain faithful over long horizons even when initialized from sparse observations. 
In GLU, this stability arises from evolving a structured latent state rather than a single compressed code or a fully unstructured token set, thereby preserving the geometry of the latent dynamics under sparse initialization.

\begin{figure}[htbp]
    \centering
    % Placeholder for image
    \includegraphics[width=\linewidth]{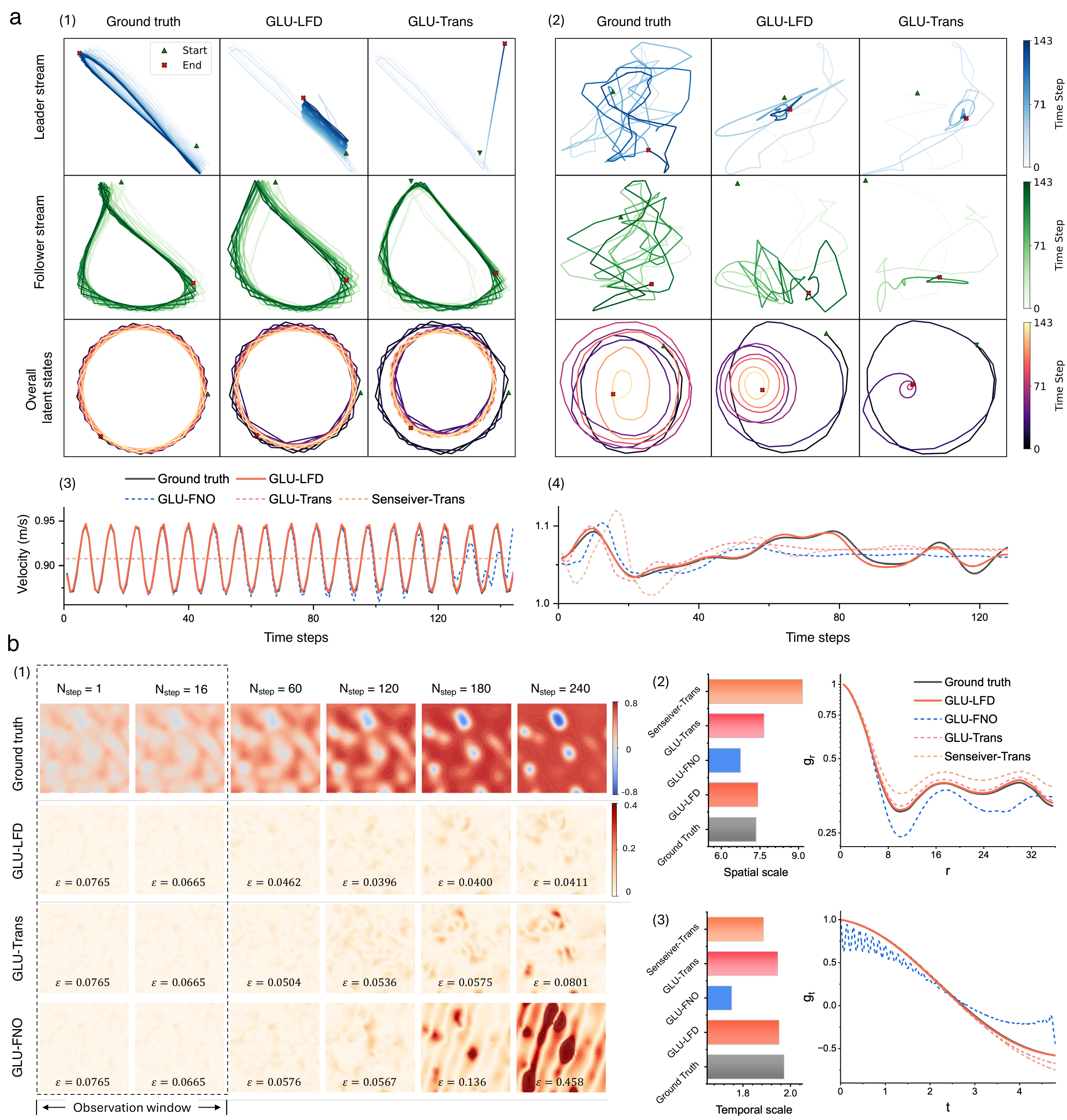}
    \caption{
    \textbf{Long-term forecasting stability on complicated dynamics.} 
    \textbf{a,} Latent dynamics visualization. 
    2D PCA projection of latent trajectories for (1) periodic (Collinear Re40, 32 sensors) and (2) chaotic (Collinear Re100, 96 sensors) flows. 
    Observation windows contain 16 steps.
    % GLU-LFD learns stable limit cycles and strange attractors that closely match the Ground Truth, whereas standard Transformer evolution (GLU-Trans) suffers from trajectory drift. 
    (3-4) Corresponding pointwise velocity forecasting confirms that GLU-LFD avoids the phase shifts and amplitude collapses seen in baselines. 
    \textbf{b,} Generalization to unseen dynamics. 
    (1) Error evolution of the reaction-diffusion system from unseen random initial condition during training from 16 steps of sparse observations (64 sensors). 
    % Error maps (bottom rows) show that baselines rapidly accumulate structural errors, while GLU-LFD maintains low error levels. 
    (2-3) Quantitative stability metrics. 
    GLU-LFD preserves the correct spatial ($g_r$) and temporal correlation ($g_t$) structures over time, preventing the physical degradation.
    }
    \label{fig:temporal_stability}
\end{figure}

We project the high-dimensional latent state into a 2D subspace through principal component analysis (PCA).
As shown in \textbf{Figure~\ref{fig:temporal_stability}a-(1) \& (3)}, for the periodic Collinear Re40 flow, the ground truth dynamics form a stable limit cycle. 
GLU-LFD preserves this closed-orbit geometry across the latent hierarchy. 
The leader stream remains confined to a smooth global trajectory, the follower stream retains the local cyclic structure, and the combined latent state recovers the correct periodic manifold. 
In contrast, standard Transformer evolution introduces progressive drift, leading to distorted loops and eventual phase error.

In the Collinear Re100 regime (\textbf{Fig.~\ref{fig:temporal_stability}a-(2),(4)}), long-horizon accuracy depends on remaining confined to the correct attractor geometry rather than matching pointwise trajectories indefinitely. 
GLU-LFD preserves a bounded latent trajectory that closely follows the topology of the ground-truth attractor, whereas transformer- and operator-based baselines drift toward distorted or collapsed trajectories. 
This geometric fidelity delays error growth in the observable dynamics and allows tracking chaotic velocity fluctuations for substantially longer horizons. 
Thus, GLU-LFD prevents the accumulation of high-frequency noise that typically destabilizes fully autoregressive models.
The comparison between GLU-LFD and GLU-Trans isolates the role of the hierarchical evolution law: both models share the same latent representation, but only LFD preserves the correct orbit geometry over long rollouts.
% The comparison of full-field error evolution of the two collinear plate flow cases is provided in the SI.

We further challenge the model's generalization capability by using the reaction-diffusion system initialized from unseen random conditions. 
As shown in \textbf{Fig.~\ref{fig:temporal_stability}b-(1)}, error maps show that baseline rollouts accumulate coherent structural errors along propagating fronts, which progressively distort the wave pattern and displace activation regions over time. 
GLU-LFD suppresses this front-linked error amplification, preserving sharper wavefronts and substantially reducing the build-up of spatial artifacts.
Quantitative metrics confirm this structural preservation. 
The spatial correlation length ($g_r$) reflects the characteristic size of coherent structures in the reaction-diffusion field, corresponding to the typical spacing and width of the propagating wavefronts generated by the interaction of reaction kinetics and diffusion. 
The evolution of the spatial correlations in \textbf{Figure~\ref{fig:temporal_stability}b-(2)} shows that GLU-LFD maintains the correct spatial correlation length scale throughout the simulation. 
In contrast, baseline errors manifest as either over-smoothed fronts or spurious small-scale structures, both of which shift the correlation length away from its physical value.
% Preserving this quantity indicates that the model reproduces the correct spatial pattern scale of the underlying dynamics. 
% In contrast, baseline models tend to distort this structural scale: over-smoothing increases the spatial correlation length and produces excessively diffusive wavefronts, whereas noisy predictions decrease the correlation length by introducing small-scale structures and fragmented activation patterns that do not exist as shown with GLU-FNO.

A similar trend appears in time. 
The temporal correlation ($g_t$) length represents the intrinsic dynamical timescale over which the system remains correlated in time and therefore reflects the characteristic propagation and relaxation times of the reaction kinetics.
GLU-LFD most closely matches the temporal correlation scale of the ground truth (\textbf{Fig.~\ref{fig:temporal_stability}b-(3)}), showing that it preserves the intrinsic timescale of wave propagation and relaxation. Baseline models decorrelate either too quickly or too slowly, corresponding to temporally jittery dynamics or artificially persistent structures respectively.
% Additional comparisons of full-field error evolution of unseen random conditions in the reaction-diffusion system are provided in the SI.
Together, these results show that GLU-LFD improves long-horizon forecasting not merely by reducing instantaneous error, but by preserving the latent geometry and correlation structure of the underlying dynamics under sparse initialization. 

\subsection*{Learning realistic system with multi-physics coupling}
\label{Subsec: RealisticDataset}

Having established the reconstruction and forecasting mechanisms of GLU on canonical systems, we then challenge GLU with a realistic multi-physics system: a turbulent premixed combustion flame. 
This system is characterized by the tight coupling between fluid dynamics (turbulence) and chemical kinetics (reactions), resulting in multi-scale intermittency and sharp gradients that typically confound sparse reconstruction methods~\cite{WANG2026114178}.

\begin{figure}[htbp]
    \centering
    \includegraphics[width=\linewidth]{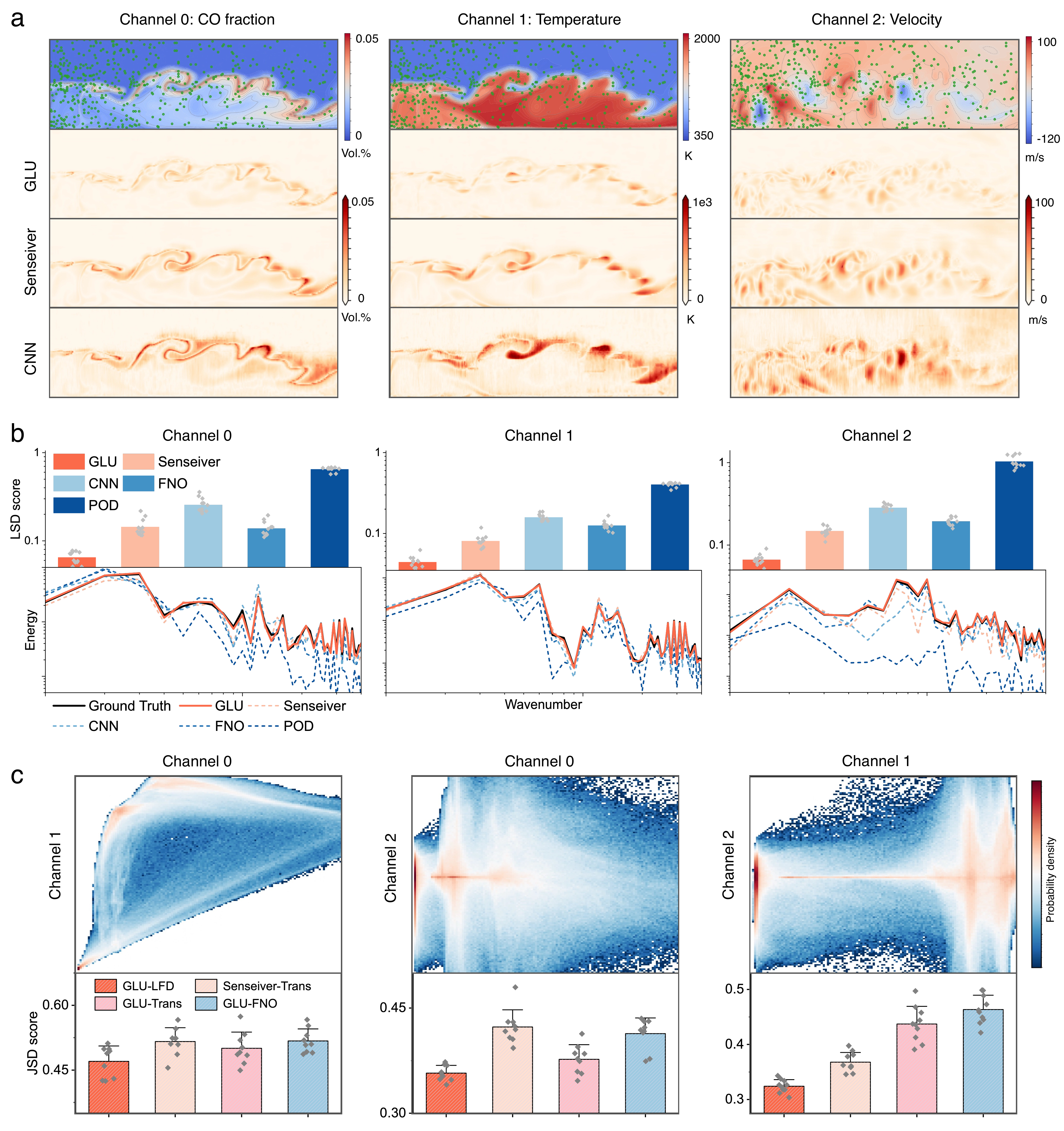}
    \caption{\textbf{Performance of GLU in learning realistic multi-physics systems (turbulent combustion).} 
    \textbf{a,} Instantaneous snapshots and error maps for all channels from 1\% sparse sensors. 
    GLU minimizes structural errors at the flame front (sharp interfaces) and in turbulent regions compared to baselines, which suffer from blurring. 
    \textbf{b,} Energy spectra (bottom) and log-spectral distance (LSD) scores (top). 
    GLU adheres to the ground truth spectrum at high wavenumbers, capturing fine-scale turbulence that baselines filter out. 
    \textbf{c,} Joint probability density functions visualizing cross-channel correlations. 
    The low Jensen-Shannon divergence (JSD) scores confirm that GLU accurately preserves the thermodynamic and kinematic couplings inherent in combustion.
    }
    \label{fig:multiphysics}
\end{figure}

\textbf{Figure~\ref{fig:multiphysics}a} presents the instantaneous reconstruction of three coupled fields from sparse sensors that cover only 1\% of the domain, including the mass fraction of carbon monoxide (CO), temperature, and velocity. 
Qualitative comparison reveals a striking difference in fidelity. 
Baseline methods (CNN, Senseiver) exhibit significant structural error patterns (red regions in error maps), particularly along the flame front where chemical reactions are most active. 
They tend to blur these sharp interfaces, a symptom of information loss in global bottlenecks or fixed-radius smoothing. 
In contrast, GLU reconstructs the flame topology with high precision while retaining the small-scale eddies in the background flow. 
This is precisely the regime that needs the global-local design: the global context maintains coherent flame geometry, while importance-weighted local retrieval concentrates resolution near the most informative regions.
% Detailed quantitative reconstruction and forecasting errors of all models in this dataset are provided in the SI.

We quantify the multi-scale fidelity via spectral analysis in \textbf{Figure~\ref{fig:multiphysics}b}. The energy spectra of the GLU reconstructions closely adhere to the ground truth (black lines) across the entire wavenumber range, effectively recovering the inertial subrange and high-frequency turbulent fluctuations that are systematically filtered out by the baselines. 
This visual observation is corroborated by the log-spectral distance (LSD), with the clearest gains at high wavenumbers where fine turbulent and interfacial structures reside. 
Defined as the root-mean-square distance between the logarithmic power spectra of the ground truth $P(\mathbf{k})$ and the reconstruction $\hat{P}(\mathbf{k})$:
\begin{equation}
\text{LSD} = \sqrt{ \frac{1}{K} \sum_{\mathbf{k}} \left| \log_{10} P(\mathbf{k}) - \log_{10} \hat{P}(\mathbf{k}) \right|^2 },
\end{equation}
this metric serves as a rigorous measure of texture preservation across scales~\cite{iotti2025rainscalegan}. 
GLU achieves the lowest LSD error across all physical channels, confirming that the fusion of local sensor tokens allows the model to bypass the spectral bias often seen in surrogate models like neural operators, where high-frequency details are suppressed.

Beyond the accuracy of individual fields, a true digital twin must preserve the underlying physical laws that govern the system.  
\textbf{Figure~\ref{fig:multiphysics}c} analyzes these cross-channel correlations via Joint Probability Density Functions (PDFs). 
For all models tested, the joint PDFs are evaluated by averaging 96 continuous snapshots in the time dimension, including 32 frames as the initial sparse observation window (with 192 random sensors), to test the physical fidelity across the long-term forecasting process.
The complex, non-linear manifolds formed by the interaction of reacting species, temperature, and velocity are accurately captured by GLU.
To quantify this topological agreement, we employ the Jensen-Shannon divergence (JSD)~\cite{NEURIPS2021_fe2d0103} to measure the similarity between the full PDFs. 
% For two PDFs from $P$ (ground truth) and $Q$ (predicted), the JSD is defined as:
% \begin{equation}
% \text{JSD}(P | Q) = \frac{1}{2} D_{\text{KL}}(P | M) + \frac{1}{2} D_{\text{KL}}(Q | M), \quad \text{where } M = \frac{1}{2}(P+Q).
% \end{equation}
% Here, $D_{\text{KL}}$ denotes the Kullback-Leibler divergence. 
% Unlike simple correlation coefficients, the JSD is a symmetric, bounded metric that measures the similarity between the full distributions. 
The JSD is a symmetric, bounded metric that measures the similarity between the full distributions. 
GLU consistently achieves the lowest JSD scores compared to baselines, which often generate statistically independent errors that fracture these correlations. 
This critical result demonstrates that GLU is not merely memorizing image patterns, but has successfully learned the underlying multi-physics manifold. 
Many reconstruction errors are physically acceptable only when viewed channel-wise: a prediction can look plausible in isolation yet still violate the cross-variable relationships that govern heat release, transport, and flame structure.
GLU thus serves as a physics-consistent inference engine, enabling reliable monitoring of multi-physics industrial systems.

% --------------------------------------------------------------------------------------

\section*{Discussion}
\label{sec: Discussion}

Digital twins are ultimately useful only if they can reliably recover the state of a system and propagate that state forward in time under realistic sensing constraints.
In this work, we framed these two functions as a single state-representation problem under partial observability. 
The central idea of GLU is that a sparse digital twin should not rely on a purely global compressed code or on purely local sensor interpolation, but instead learn a state that is globally predictive, locally faithful, and adaptively weighted by uncertainty. 
The value of GLU therefore lies not only in a new architecture, but in a representation design that makes reconstruction, forecasting, and sparse information integration part of the same computational object.
GLU improves not only pointwise accuracy, but also multi-scale reconstruction fidelity, long-horizon stability, and physically meaningful cross-channel consistency. 
More broadly, this framing is aligned with how digital twins are increasingly viewed in process engineering and industrial ML, where real-time monitoring, prediction, optimization, and control depend on scalable data assimilation rather than isolated reconstruction or forecasting modules alone~\cite{peterson2025digital}.

\begin{figure}[htbp]
    \centering
    % Placeholder for image
    \includegraphics[width=\linewidth]{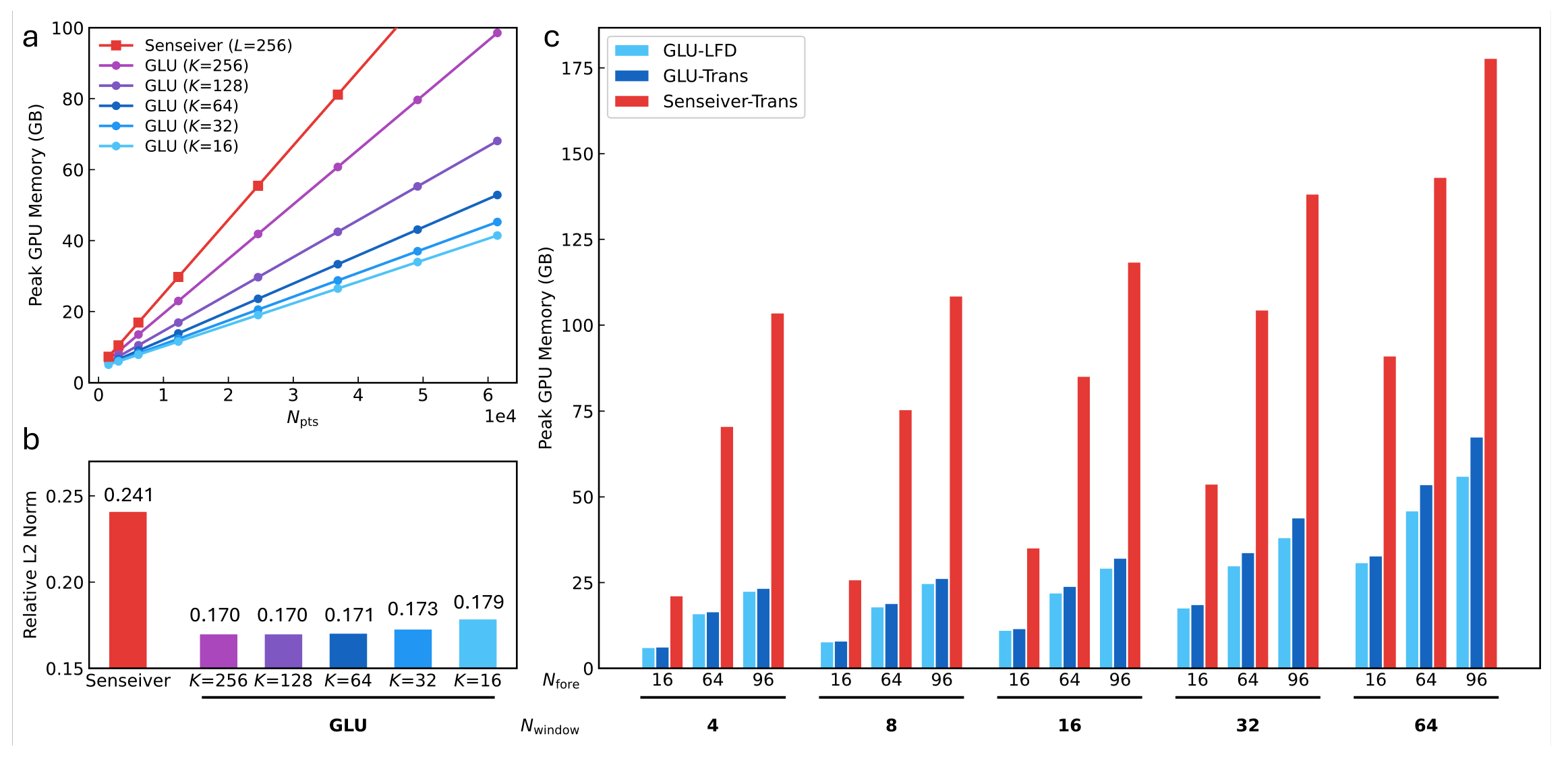}
    \caption{
    \textbf{Scalability and memory comparison for spatial and temporal reconstruction.} 
    \textbf{a,} Peak GPU memory versus the number of reconstruction points, $N_{\mathrm{pts}}$, for Senseiver and GLU with different information-bottleneck top-$K$ sensor settings. 
    Lower top-$K$ reduces GLU memory, enabling a tunable efficiency--accuracy trade-off while maintaining substantially better memory scaling than Senseiver. 
    \textbf{b,} Corresponding relative $L_2$ norm for the models on turbulent channel flow dataset. 
    % Although error increases slightly as top-$K$ decreases, all GLU variants remain more accurate than Senseiver. 
    \textbf{c,} Peak GPU memory for temporal reconstruction as a function of the number of reference frames, $N_{\mathrm{ref}}$, and window size, $N_{\mathrm{window}}$. 
    GLU-LFD achieves the lowest memory usage across all temporal settings.
    % , outperforming both GLU-Trans and Senseiver-Trans.
    }
    \label{fig:scalability}
\end{figure}

A practical deployment is constrained not only by accuracy but also by computational cost. 
The scalability tests in \textbf{Figure~\ref{fig:scalability}} show that this trade-off is favorable for GLU. 
In reconstruction, the peak GPU memory of GLU grows substantially more slowly with the number of query points compared to Senseiver. This advantage is retained even when the adaptive reconstructor uses increasingly large top-$K$ neighborhoods, as indicated in \textbf{Figure~\ref{fig:scalability}(a)}. 
Importantly, \textbf{Figure~\ref{fig:scalability}(b)} further shows that the smallest tested GLU setting ($K$=16) already surpasses Senseiver in reconstruction accuracy while using markedly less memory, indicating that the structured latent decomposition is not only more expressive, but also more efficient in how it allocates memory and computation across sensing and query scales. 

This advantage extends naturally to temporal forecasting, where model cost must scale with both the number of reference frames and the length of the observation window, as shown in \textbf{Figure~\ref{fig:scalability}(c)}.
GLU-LFD is consistently more memory efficient, while Senseiver-Trans becomes particularly costly as both the observation window and rollout horizon increase. 
By evolving the latent state through the hierarchical Leader--Follower--Dynamics mechanism, GLU avoids the dense token--token interactions used by standard transformer evolution and instead concentrates computation on the most informative information pathways.
Thus, imposing structure on the latent state is valuable not only for stability, but also for scaling digital twins to higher reconstruction resolution and longer forecasting horizons, both of which are central bottlenecks in real deployments.
GLU is also complementary to data assimilation. 
Classical and neural DA methods focus on how observations should refine a state estimation or update a forecast model~\cite{Gou2024Global}, whereas GLU focuses on how sparse observations should be assembled into a structured state representation that is both inferable and forecastable. 
A natural next step is therefore to combine the two, using GLU as a learned state representation or observation operator within sequential or variational assimilation pipelines.

More broadly, the design rationale of GLU is not specific to deterministic prediction. 
A structured latent state that separates global organization, local measurement fidelity, and uncertainty-aware information weighting should also provide a natural conditioning interface for diffusion- or flow-matching-based generative models.
In this setting, the role of GLU would be to impose representation structure and scalability, while the generative process would provide a richer posterior uncertainty and future evolution~\cite{zhuang2025spatially}.
At the application-level, GLU provides a promising backbone for field-aware industrial digital twins that offers a unified route from sparse measurement ingestion to full-state estimation and dynamic forecasting, while retaining a representation that is sufficiently structured to support downstream tasks such as sensor adaptation, anomaly detection, and eventually closed-loop control~\cite{lawrence2024machine}. 

\section*{Methods}
\label{sec: Methods}

\subsection*{Dataset preparations}
\label{subsec: Datasets}

We considered seven datasets spanning organized periodic flows, chaotic turbulence, geophysical fields, nonlinear reaction--diffusion dynamics and realistic coupled multi-physics combustion.
To provide a unified learning setting across all benchmarks, we reformatted every dataset into a field tensor of shape $[n_{\mathrm{cases}},\, n_t,\, n_p,\, n_c]$ together with a coordinate array of shape $[n_p,\, n_d]$, where $n_{\mathrm{cases}}$ denotes the number of independent cases or trajectories, $n_t$ the number of time steps, $n_p$ the number of spatial points, $n_c$ the number of physical channels, and $n_d$ the spatial dimension. 
All field variables were standardized by mean--variance normalization, and all spatial coordinates were linearly rescaled to $[-1,1]$. 
Unless otherwise noted, $85\%$ of the cases were used for training and the remaining $15\%$ for testing. 
During training, the data-fitting term was the mean-squared reconstruction error. 

\bmhead{Cylinder flow}
We first considered the two-dimensional laminar cylinder flow at $Re_D=100$, following Fukami \emph{et al.}~\cite{fukami2021global}. 
The computational domain covers $(x/D,y/D)\in[-0.7,15]\times[-5,5]$ with spatial resolution $192\times112$, and the vorticity field is used as the single output channel. This dataset contains coherent vortex shedding and well-defined near- and far-wake structures, and was selected as a canonical benchmark for sparse reconstruction of organized unsteady flow fields.

\bmhead{NOAA sea surface temperature}
We next used the NOAA sea surface temperature dataset provided by Fukami \emph{et al.}~\cite{fukami2021global}, consisting of weekly observations on a $360\times180$ global grid. 
Following the source study, the dataset contains 1,040 snapshots spanning 1981--2001 for model development, with later snapshots from 2001--2018 used for generalization assessment. 
We included this benchmark to test the model performance on a large-scale geophysical field with coastlines, transport-dominated variability, and strongly non-uniform spatial informativeness.

\bmhead{Turbulent channel flow}
To probe broadband chaotic dynamics, we used the turbulent channel-flow dataset provided by Fukami \emph{et al.}~\cite{fukami2021global} at friction Reynolds number $Re_{\tau}=180$. 
The original direct numerical simulation domain is $(L_x,L_y,L_z)=(4\pi\delta,2\delta,2\pi\delta)$ with grid size $(256,96,256)$. 
Following the source treatment, we extracted an $x$--$y$ subdomain over $[0,2\pi\delta]\times[0,\delta]$ with spatial resolution $128\times48$, and used fluctuating streamwise velocity $u'$ as the target variable. 
This dataset contains multi-scale turbulent structures with weak temporal regularity, making it a demanding test of fine-scale spatial recovery from sparse observations.

\bmhead{Collinear plate flow}
For forecasting experiments, we used the two-dimensional collinear flat-plate wake dataset of Solera-Rico \emph{et al.}~\cite{SoleraRico2024} at $Re=40$ and $Re=100$. 
The source simulations were performed using an immersed-boundary projection method on nested grids and subsequently downsampled to a uniform spatial resolution of $300\times98$. 
The fluctuating streamwise and cross-stream velocity components were stacked as two channels, and we mainly tested our model on the velocity in the $x$--direction. 
The $Re=40$ case exhibits approximately periodic wake evolution, whereas the $Re=100$ case lies in a substantially more chaotic regime.
Together, they provide a controlled test of long-horizon forecasting under both regular and irregular attractor structures.

\bmhead{FitzHugh--Nagumo reaction--diffusion system}
We further generated an in-house two-dimensional dataset based on the FitzHugh--Nagumo  reaction--diffusion system
\begin{equation}
u_t=\mu_u\Delta u + u-u^3-v+\alpha,\qquad
v_t=\mu_v\Delta v + \beta (u-v),
\end{equation}
following the formulation in Ref.~\cite{santos2023development}. 
Simulations were performed on $\Omega=[-50,50]^2$ with $\Delta x=1.0$, $\Delta t=0.02$, periodic boundary conditions and Gaussian-noise initial conditions with mean $0$ and standard deviation $0.05$. 
Each sample contains two channels corresponding to the state variables $u$ and $v$, and we mainly tested the model on the $u$ channel. 
This dataset introduces nonlinear pattern formation and propagating wavefronts from random initial conditions, providing a complementary benchmark for sparse reconstruction and forecasting in systems governed by reaction and diffusion rather than advection alone.

\bmhead{Turbulent combustion}
Finally, we considered a realistic multi-physics dataset generated in-house from a large-eddy simulation of turbulent methane--air combustion over a two-dimensional backward-facing step using \texttt{reactingFoam} in \texttt{OpenFOAM}~\cite{Jasak2009OpenFOAM}. 
% The simulation employed the one-equation eddy-viscosity subgrid model (\texttt{kEqn}), finite-rate chemistry with the GRI-3.0 mechanism, and radiative heat transfer via the P-1 model. 
A stoichiometric methane--air mixture entered the domain at 20~m$\cdot$s$^{-1}$ with imposed turbulent velocity fluctuations and an inlet temperature of 400~K. 
The flame was stabilized behind the step using an initialized ignition kernel. 
Snapshots were saved every $5\times10^{-5}$~s over a total duration of 0.5~s, in total 10,000 snapshots. 
We retained three channels, including the mole fraction of CO, temperature, and streamwise velocity, to construct the dataset. 
This benchmark is the most challenging test case that combines thin reaction fronts, broadband turbulent transport, and strong cross-channel coupling, thereby approximating the sensing conditions encountered in practical digital twins of reactive flow systems.

\subsection*{Global-local-aware encoder}
\label{subsec: encoder}

The global-local-aware encoder transforms sparse, unstructured sensor measurements into a unified latent representation without information loss. 
Let the input at time $t$ be a set of $N$ measurements $\mathcal{X}_t = \{(\mathbf{x}_i, \mathbf{u}_i)\}_{i=1}^N$, where $\mathbf{x}_i \in \mathbb{R}^d$ denotes the spatial coordinates and $\mathbf{u}_i \in \mathbb{R}^c$ denotes the physical quantities.
We first map the low-dimensional coordinates and values into a high-dimensional feature space.
To capture high-frequency spatial variations, we employ Fourier feature mapping $\gamma(\cdot)$:
\begin{equation}
    \mathbf{e}_i^{\text{pos}} = \mathbf{W}_p \cdot [\cos(2\pi \mathbf{B} \mathbf{x}_i), \sin(2\pi \mathbf{B} \mathbf{x}_i)], \quad \mathbf{e}_i^{\text{val}} = \mathbf{W}_v \cdot \mathbf{u}_i
\end{equation}
where $\mathbf{B}$ is a learnable frequency matrix initialized from a Gaussian distribution, and $\mathbf{W}_p, \mathbf{W}_v$ are linear projections. 
The initial sensor embedding is obtained via element-wise addition: $\mathbf{h}_i = \text{LayerNorm}(\mathbf{e}_i^{\text{pos}}) + \text{LayerNorm}(\mathbf{e}_i^{\text{val}})$.

Our architecture enforces a bidirectional information flow to preserve local fidelity. 
We initialize a set of $S$ learnable latent queries $\mathbf{L} \in \mathbb{R}^{S \times D}$. The encoding proceeds in two stages.
In the global aggregation stage, the latent queries $\mathbf{L}$ attend to the sensor embeddings $\mathbf{H} = \{\mathbf{h}_i\}$ to extract a compact system-level summary:
\begin{equation}
    \mathbf{L}' = \text{CrossAttention}(\mathbf{Q}=\mathbf{L}, \mathbf{K}=\mathbf{H}, \mathbf{V}=\mathbf{H})
\end{equation}
This operation scales as $\mathcal{O}(N \cdot S)$, efficiently aggregating information from arbitrary sensor counts.
Then, in the local enrichment stage, the refined global context is broadcast back to the original sensor tokens.
The sensor embeddings $\mathbf{H}$ query the updated latent array $\mathbf{L}'$:
    \begin{equation}
        \mathbf{Z}_{\text{local}} = \text{CrossAttention}(\mathbf{Q}=\mathbf{H}, \mathbf{K}=\mathbf{L}', \mathbf{V}=\mathbf{L}') + \mathbf{H}
    \end{equation}
% This residual connection ensures that the specific local measurement of every sensor is preserved.
We select the last token in $\mathbf{L}$ as the global summary $\mathbf{z}_{\text{global}}$.
% The final output is the set union of the enriched sensor tokens and the global leader token: $\{ \mathbf{Z}_{\text{global}}, \mathbf{Z}_{\text{local}} \}$.

\subsection*{Soft domain-adaptive reconstructor}
\label{subsec: Reconstructor}

The soft domain-adaptive reconstructor recovers the continuous physical field from the discrete latent representation.
This module determines the value at an arbitrary query location $\mathbf{y}$ by aggregating information from its most relevant neighbors in a learnable metric space.

Let $\phi_i \in (0, 1)$ be the learned importance score for the $i$-th sensor. 
We define an asymmetric, adaptive distance metric $d_{\phi}$ between a query $\mathbf{y}$ and a sensor $\mathbf{x}_i$:
\begin{equation}
    d_{\phi}(\mathbf{y}, \mathbf{x}_i) = \frac{\| \mathbf{y} - \mathbf{x}_i \|_2}{\phi_i^\gamma + \epsilon}
\end{equation}
where $\gamma$ is a scaling factor and $\epsilon$ is for numerical stability. Sensors with high importance scores ($\phi_i \approx 1$) effectively shrink their distance to the query, expanding their region of influence, while unimportant sensors are pushed away.

Then, for each query point $\mathbf{y}_j$, we identify the set $\mathcal{N}_j$ of the $k$ nearest sensors by the metric $d_{\phi}$. The local feature $\mathbf{f}_j$ is computed via a soft-weighted aggregation:
\begin{equation}
    w_{ij} = \frac{\exp(-d_{\phi}(\mathbf{y}_j, \mathbf{x}_i) / \sigma_i)}{\sum_{m \in \mathcal{N}_j} \exp(-d_{\phi}(\mathbf{y}_j, \mathbf{x}_m) / \sigma_m)},
\end{equation}
\begin{equation}
    \mathbf{f}_j = \sum_{i \in \mathcal{N}_j} (w_{ij} \cdot \phi_i) \cdot \text{Proj}(\mathbf{z}_i),
\end{equation}
where $\sigma_i$ is a bandwidth parameter set to 0.05, and $\mathbf{z}_i$ denotes the $i$-th row of $\mathbf{Z}_{\text{local}}$. 
% This formulation is fully differentiable with respect to the feature space.
After neighborhood selection, the local aggregation is differentiable with respect to the token features and fusion parameters.

To ensure global consistency, the aggregated local feature $\mathbf{f}_j$ is fused with the global leader token $\mathbf{z}_{\text{global}}$ and the query's positional embedding $\mathbf{p}(\mathbf{y}_j)$:
\begin{equation}
    \mathbf{h}_j = \text{MLP}_{\text{fusion}}([\mathbf{f}_j + \mathbf{p}(\mathbf{y}_j); \mathbf{z}_{\text{global}}])
\end{equation}
Finally, two separate decoding heads project $\mathbf{h}_j$ to the predicted mean field value $\hat{u}(\mathbf{y}_j)$ and the aleatoric reconstruction uncertainty $\hat{\sigma}^2(\mathbf{y}_j)$.

\subsection*{Variational learning of uncertainty-driven importance scores}
\label{subsec:ImportanceScore}

We model the spatial importance score as a bounded latent variable \(\phi(\mathbf{x}) \in (0,1)\) defined over spatial coordinates. Rather than prescribing this score heuristically, we learn it from the model's own reconstruction uncertainty so that regions that are consistently difficult to infer under sparse observations receive higher importance.
For each coordinate \(\mathbf{x}\), a neural network predicts the parameters of a Beta distribution~\cite{NEURIPS2018_92c8c96e},
\begin{equation}
q_{\psi}\!\left(\phi(\mathbf{x}) \mid \mathbf{x}\right)
=
\mathrm{Beta}\!\left(\alpha(\mathbf{x}), \beta(\mathbf{x})\right),
\end{equation}
with
\begin{equation}
\alpha(\mathbf{x}) = \exp\!\big(f_{\psi}^{(\alpha)}(\mathbf{x})\big) + \varepsilon,
\qquad
\beta(\mathbf{x}) = \exp\!\big(f_{\psi}^{(\beta)}(\mathbf{x})\big) + \varepsilon,
\end{equation}
where \(\varepsilon>0\) ensures numerical stability. The posterior mean
\begin{equation}
\bar{\phi}(\mathbf{x}) = \frac{\alpha(\mathbf{x})}{\alpha(\mathbf{x})+\beta(\mathbf{x})}
\end{equation}
is used as the deterministic importance score during decoding and adaptive sensor selection.

% For forecasting experiments, we optionally introduce a second Beta factor,
% \begin{equation}
% q_{\psi_2}\!\left(\phi_2(\mathbf{x}) \mid \mathbf{x}\right)
% =
% \mathrm{Beta}\!\left(\alpha_2(\mathbf{x}), \beta_2(\mathbf{x})\right),
% \end{equation}
% and define the final importance score as the product of posterior means,
% \begin{equation}
% \bar{\phi}(\mathbf{x}) = \bar{\phi}_1(\mathbf{x})\,\bar{\phi}_2(\mathbf{x}).
% \end{equation}
% In reconstruction-only settings, \(\bar{\phi}_2(\mathbf{x}) \equiv 1\).

The supervisory signal for importance learning is derived from the variance head of the soft domain-adaptive reconstructor. In addition to the reconstructed mean field \(\hat{u}(\mathbf{x})\), the reconstructor predicts a log-variance \(\log \hat{\sigma}^2(\mathbf{x})\), from which we form a normalized uncertainty target
\begin{equation}
U(\mathbf{x}) = \mathrm{Norm}\!\left(\hat{\sigma}^2(\mathbf{x})\right),
\end{equation}
where \(\mathrm{Norm}(\cdot)\) denotes min--max normalization in the spatial domain. 
% During forecasting training, this spatial uncertainty map can be optionally blended with a global temporal-uncertainty term estimated from the latent log-variance,
% \begin{equation}
% U(\mathbf{x}) = (1-\lambda_t)\,\mathrm{Norm}\!\left(\hat{\sigma}^2(\mathbf{x})\right) + \lambda_t\,\tau,
% \end{equation}
% where \(\tau\) is the normalized mean latent variance and \(\lambda_t\in[0,1]\) is an annealed blending weight.
We optimize the importance model jointly with the main training objective using an ELBO-style auxiliary loss~\cite{Kingma2014},
\begin{equation}
\mathcal{L}_{\phi}
=
- \mathbb{E}_{q_{\psi}}\!\left[
\frac{1}{|\Omega|}\sum_{\mathbf{x}\in\Omega} U(\mathbf{x})\,\phi(\mathbf{x})
\right]
+ \lambda_{\mathrm{KL}}\,D_{\mathrm{KL}}\!\left(q_{\psi}(\phi \mid \mathbf{x}) \,\|\, p(\phi)\right)
- \lambda_{\mathrm{H}}\,\mathcal{H}\!\left[q_{\psi}(\phi\mid \mathbf{x})\right]
- \lambda_{\mathrm{V}}\,\mathrm{Var}_{\mathbf{x}}\!\left[\bar{\phi}(\mathbf{x})\right].
\end{equation}
Here, \(p(\phi)=\mathrm{Beta}(\alpha_0,\beta_0)\) is a Beta prior, \(\mathcal{H}[\cdot]\) denotes differential entropy, and \(\mathrm{Var}_{\mathbf{x}}[\bar{\phi}(\mathbf{x})]\) encourages a non-degenerate spatial distribution of importance scores. The KL and entropy terms are applied to the factor updated in the current training stage. The expectation term is estimated by Monte Carlo sampling from the Beta distributions using the reparameterization trick.

This objective encourages high importance in regions with persistently elevated predictive uncertainty, while regularizing the learned importance field toward a stable prior~\cite{margossian2024amortized}. 
In practice, the resulting \(\bar{\phi}(\mathbf{x})\) acts as an adaptive spatial weighting function that modulates both neighborhood selection in the reconstructor.
% and, when enabled, importance-aware temporal evolution.

\subsection*{Leader–Follower-Dynamics (LFD) in latent space}
\label{subsec: Propagator}

We model latent evolution as a learned continuous-time derivative integrated with an explicit Euler step:
\begin{equation}
    \mathcal{S}_{t+1} = \mathcal{S}_t + \Delta t \cdot \mathcal{F}_\theta(\text{Pool}(\mathcal{S}_{t-w:t})),
\end{equation}
where $\mathcal{F}_\theta$ is the temporal derivative function parameterized by a stack of hierarchical Transformer blocks, and $\text{Pool}(\cdot)$ applies temporal pooling to compress the history window $w$.
To scale to large sensor sets, we employ a hierarchical Leader-Follower architecture that avoids quadratic attention complexity.

The derivative function $\mathcal{F}_\theta$ splits the processing into two asymmetric streams.
In the leader stream, the global token updates its state by aggregating information from the local sensors and its own history. 
This aggregation is gated by the importance scores $\boldsymbol{\phi}$:
\begin{equation}
    \mathbf{z}_{\text{global}}' = \text{SelfAttn}(\mathbf{G}_{t-w;t}) + \text{CrossAttn}(\mathbf{Q}=\mathbf{G}_{t-w;t}, \mathbf{KV}=\mathbf{Z}_{\text{local}}, \text{bias}=\ln \boldsymbol{\phi}),
\end{equation}
where $\mathbf{G}_{t-w;t}=[\mathbf{z}_{\text{global}}^{(t-w)},\dots,\mathbf{z}_{\text{global}}^{(t)}]$ denotes the available history of global token.
The term $\ln \boldsymbol{\phi}$ acts as an additive mask on the attention logits, suppressing contributions from unimportant or unreliable sensors.
    
Then, in the follower stream, the sensor tokens evolve by attending solely to the leader, downloading global physics updates without direct peer-to-peer communication:
\begin{equation}
    \mathbf{z}_{i}' = \text{CrossAttn}(\mathbf{Q}=\mathbf{z}_i, \mathbf{KV}=\mathbf{G}_{t-w;t}).
\end{equation}
For fixed history length and leader width, this reduces the dominant sensor-scaling cost of the dynamics module to linear in $N$.

\subsection*{Spatial reconstruction baseline methods}
\label{subsec: Spatial_Baseline}

We compared GLU against four representative classes of sparse field reconstruction methods spanning classical reduced-order modeling, convolutional reconstruction, neural operators, and attention-based set encoders.
For all reconstruction baselines, we strictly followed the model configurations and training settings provided in the original code-base or source papers wherever available, and modified only the input/output interfaces as needed to accommodate the unified data representation adopted in this work, namely field tensors of shape $[n_{\mathrm{cases}}, n_t, n_p, n_c]$ together with coordinate arrays of shape $[n_p, n_d]$. 

\bmhead{POD-GPR}
Proper Orthogonal Decomposition with Gaussian Process Regression (POD-GPR) was included as a classical reduced-order baseline. 
POD is first used to extract a low-dimensional basis from training snapshots, after which GPR maps sparse sensor observations to the corresponding modal coefficients, and the full field is reconstructed from the predicted coefficients and POD modes. 
POD-GPR represents a strong and widely used reduced-order approach that does not rely on deep learning for global field estimation from sparse measurements.~\cite{procacci2023adaptive}.

\bmhead{MLP-CNN}
MLP-CNN was adapted from the differentiable embedding strategy introduced as an alternative to Voronoi-based CNN reconstruction by Fukami \emph{et al.}~\cite{fukami2021global}. 
Sparse observations are first embedded onto a structured latent grid through a multilayer perceptron, after which stacked convolutional layers recover the full field. 
This baseline is a representative convolutional approach for sparse-to-dense reconstruction and provides a strong comparison against grid-based deep reconstruction models.

\bmhead{RecFNO}
RecFNO was implemented following Zhao \emph{et al.}~\cite{zhao2024recfno}. 
This model combines sparse-input embedding with Fourier Neural Operator layers, allowing reconstruction to be performed through global spectral convolutions in Fourier space. 
We chose RecFNO as a reconstruction-tailored neural operator baseline, which is also well suited for assessing how GLU compares with globally expressive operator-learning models.

\bmhead{Senseiver}
Senseiver was implemented following Santos \emph{et al.}~\cite{santos2023development} that used an encoder--decoder attention architecture. 
Sparse observations and sensor coordinates are encoded into a latent representation through cross-attention, and queried coordinates are subsequently decoded through coordinate-aware attention. 
We selected Senseiver as one of the most relevant recent baselines for reconstruction from arbitrarily sized sets of sparse observations, making it the strongest architectural comparison to the set-based design of GLU.

\subsection*{Temporal forecasting baseline methods}
\label{subsec: Temporal_Baseline}
To evaluate the proposed LFD module for latent-space evolution, we compared it against two widely used alternatives for autoregressive forecasting.

Causal Transformer (Trans) was adopted as the primary latent-space forecasting baseline, following the general temporal-transformer~\cite{SoleraRico2024}. 
In this approach, a fixed observation window of latent states is processed by stacked causal attention blocks to predict the next latent state autoregressively. 
Causal transformer is among the most common sequence models for latent dynamics learning and provides the most direct comparison to the proposed LFD architecture.
In our implementation, the transformer was adapted to evolve latent arrays rather than a single latent vector by flattening the batch and latent-token dimensions during temporal propagation. 

FNO was used as a full-state neural-operator baseline~\cite{li2021fourier}. 
Unlike LFD and the causal transformer, which propagate compact latent representations and then decode them back to the physical space, FNO directly advances the reconstructed full field in time using spectral convolution in Fourier space~\cite{Fu2025NO}, thus it functions as a complementary forecasting strategy: whether temporal dynamics are better learned on a compact latent representation or on an expanded full-state representation that may already contain reconstruction errors.

These propagators were combined with reconstruction backbones to form complete sparse-observation forecasting pipelines. 
To limit the comparison to the most competitive and practically relevant combinations, we evaluated GLU-LFD, GLU-Trans, GLU-FNO, Senseiver-Trans, and Senseiver-FNO. 
Senseiver was chosen as the reconstruction baseline in these hybrid pipelines because it was the strongest non-GLU reconstruction model in our experiments.
% , whereas a Senseiver-LFD variant is not naturally defined because LFD relies on the explicit leader--follower token hierarchy of GLU. 
For all forecasting experiments, the models were conditioned on 16 observed frames with sparse inputs and then rolled out autoregressively to assess long-horizon stability.

% \bmhead{Acknowledgments}

% The authors acknowledge the financial support from ExxonMobil through the MIT Energy Initiative as a Founding Member, MIT Sea Grant, and National Science Foundation under Grant No. CBET-2143625. J.C. acknowledges the financial support from the Jack Kent Cooke Foundation.

\bmhead{Author contribution}
L.W.: conceptualization, methodology, investigation, visualization, writing—original draft.
J.C.: methodology, investigation, visualization, writing—original draft.
N.T.: software, investigation.
Z.C.: methodology, writing—review and editing.
S.D.: supervision, writing—review and editing.

% \bmhead{Supplementary information}

% The online version contains supplementary material available at [...].

% \bmhead{Data availability}

% [\dots]

\bmhead{Code availability}

Code will be released upon publication.

\bmhead{Conflict of interest}
The authors declare no competing interests.


% \bibliography{sn-bibliography}% common bib file
% \input output.bbl

\end{document}